\documentclass[10pt,twocolumn,letterpaper]{article}

\usepackage{cvpr}
\usepackage{times}
\usepackage{epsfig}
\usepackage{graphicx}
\usepackage{amsmath}
\usepackage{amssymb}
\usepackage{subcaption}
\usepackage{booktabs} 
\usepackage{xcolor}

\usepackage[breaklinks=true,bookmarks=false]{hyperref}
\hypersetup{
    colorlinks,
    linkcolor={red!90!black},
    citecolor={green!90!black},
    urlcolor={blue!90!black}
}

\cvprfinalcopy 

\def\cvprPaperID{306} 

\ifcvprfinal\fi
\begin{document}

\title{Look into Person: Self-supervised Structure-sensitive Learning and A New Benchmark for Human Parsing}

\author{Ke Gong$^1$, \quad Xiaodan Liang$^{1,2}$, \quad Dongyu Zhang$^1$\thanks{The first two authors contribute equally to this paper. Corresponding author is Dongyu Zhang.   This work was supported by the National Natural Science Foundation of China under Grant 61401125 and 61671182.}, \quad Xiaohui Shen$^{3}$, \quad Liang Lin$^{1,4}$\\
$^1$Sun Yat-sen University \quad $^2$Carnegie Mellon University \quad $^3$Adobe Research \quad $^4$SenseTime Group (Limited) \\
	{\tt\footnotesize gongk3@mail2.sysu.edu.cn}, {\tt\small xiaodan1@cs.cmu.edu}, {\tt\small \{zhangdy27,linlng\}@mail.sysu.edu.cn}, {\tt\small xshen@adobe.com}
}

\maketitle

\begin{abstract}
Human parsing has recently attracted a lot of research interests due to its huge application potentials. However existing datasets have limited number of images and annotations, and lack the variety of human appearances and the coverage of challenging cases in unconstrained environment. In this paper, we introduce a new benchmark\footnote {The dataset is available at \url{http://hcp.sysu.edu.cn/lip}} ``Look into Person (LIP)" that makes a significant advance in terms of scalability, diversity and difficulty, a contribution that we feel is crucial for future developments in human-centric analysis. This comprehensive dataset contains over 50,000 elaborately  annotated images with 19 semantic part labels, which are captured from a wider range of viewpoints, occlusions and background complexity. Given these rich annotations we perform detailed analyses of the leading human parsing approaches, gaining insights into the success and failures of these methods. Furthermore, in contrast to the existing efforts on improving the feature discriminative capability, we solve human parsing by exploring a novel self-supervised structure-sensitive learning approach, which imposes human pose structures into parsing results without resorting to extra supervision (i.e., no need for specifically labeling human joints in model training). Our self-supervised learning framework can be injected into any advanced neural networks to help incorporate rich high-level knowledge regarding human joints from a global perspective  and improve the parsing results. Extensive evaluations on our LIP and the public PASCAL-Person-Part dataset demonstrate the superiority of our method.
\end{abstract}

\begin{figure*}[t]
\begin{subfigure}{1.0\textwidth}
   \centering
   \includegraphics[width=1\linewidth]{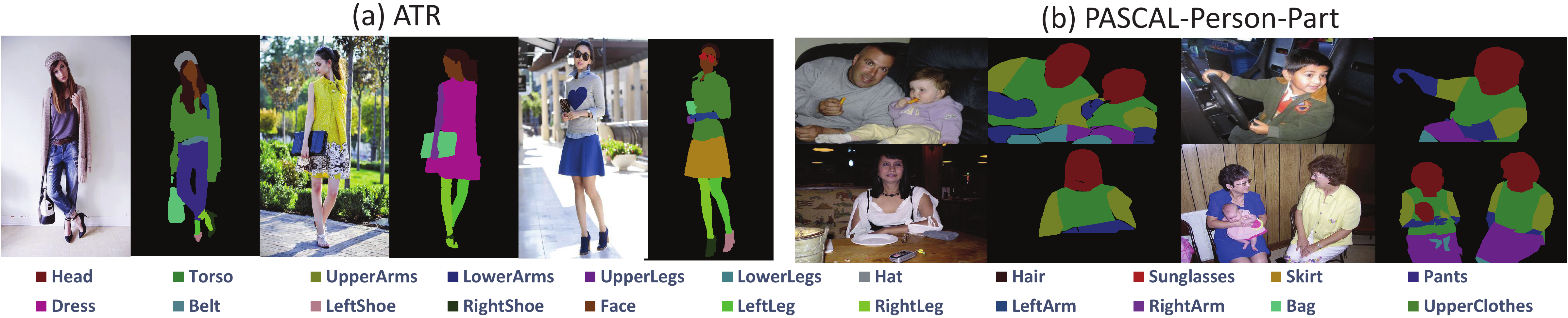}
\end{subfigure}
   \vspace{-1mm}
\begin{subfigure}{1.0\textwidth}
   \centering
   \includegraphics[width=1\linewidth]{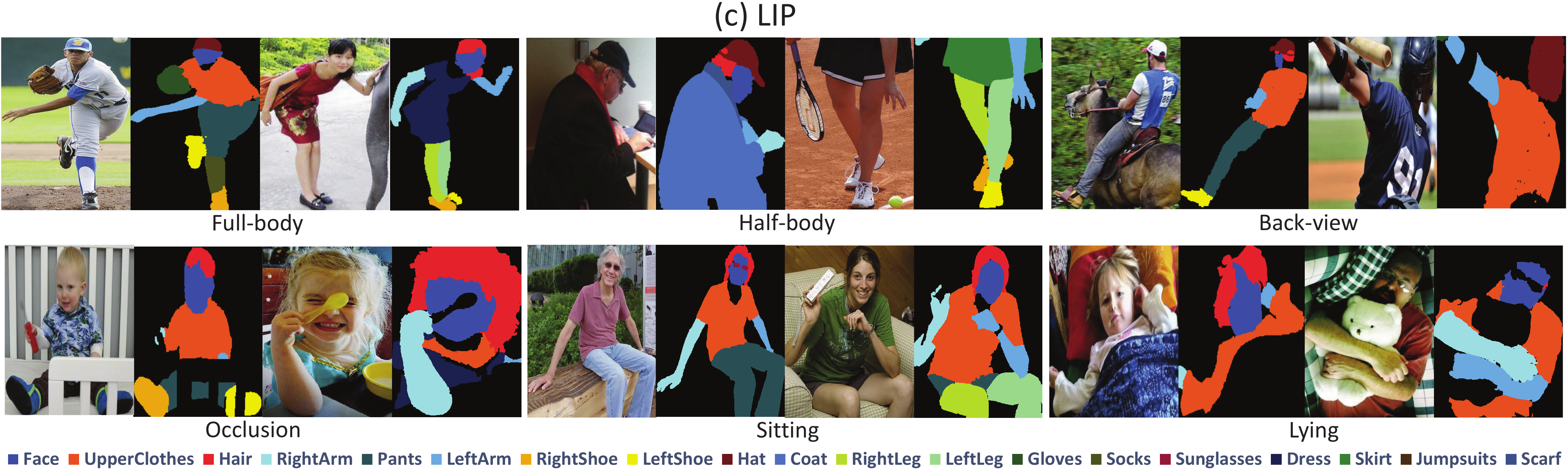}
\end{subfigure}
\vspace{-2mm}
\caption{Annotation examples for our ``Look into Person (LIP)'' dataset and existing datasets. (a) The images in ATR dataset which are fixed in size and only contain stand-up person instances in the outdoors. (b) The images in PASCAL-Person-Part dataset which also have lower scalability and only contain 6 coarse labels. (c) The images in our LIP dataset with high appearance variability and complexity.}
\vspace{-6mm}
\label{fig:dataset_example}
\end{figure*}

\section{Introduction}
Human parsing aims to segment a human image into multiple parts with fine-grained semantics and provide more detailed understanding of image contents. It can stimulate many higher-level computer vision applications~\cite{zhang2015dynamic}, such as person re-identification~\cite{zhao2013unsupervised} and human behavior analysis~\cite{gan2016concepts,liang2015proposal}. 

Recently, Convolutional Neural Networks (CNNs) have achieved exciting success in human parsing~\cite{ATR,Co-CNN,liang2015semantic}. Nevertheless, as demonstrated in many other problems such as object detection~\cite{liang2015towards} and semantic segmentation~\cite{crfasrnn}, the performance of those CNN-based approaches heavily rely on the availability of annotated images for training. In order to train a human parsing network with potentially practical value in real-word applications, it is highly desired to have a large-scale dataset composed of representative instances with varied clothing appearances, strong articulation, partial (self-)occlusions, truncation at image borders, diverse viewpoints and background clutters. Although there exist training sets for special scenarios such as fashion pictures~\cite{Yamaguchiparsing13,Dongparsing13,ATR,Co-CNN} and people in constrained situations (e.g., upright)~\cite{chen2014detect}, these datasets are limited in their coverage and scalability, as shown in Fig.~\ref{fig:dataset_example}. The largest public human parsing dataset~\cite{Co-CNN} so far only contains 17,000 fashion images while others only include thousands of images. 

Moreover, to the best of our knowledge, no attempt has been made to establish a standard representative benchmark aiming to cover a wide pallet of challenges for the human parsing task. The existing datasets did not provide an evaluation server with a secret test set to avoid potential dataset over-fitting, which hinders further development on this topic. Therefore we propose a new benchmark ``Look into Person (LIP)" and a public server for automatically reporting evaluation results. Our benchmark significantly advances the state-of-the-arts in terms of appearance variability and complexity, which includes 50,462 human images with pixel-wise annotations of 19 semantic parts.

\begin{figure}[]
\centering
   \includegraphics[width=0.8\linewidth]{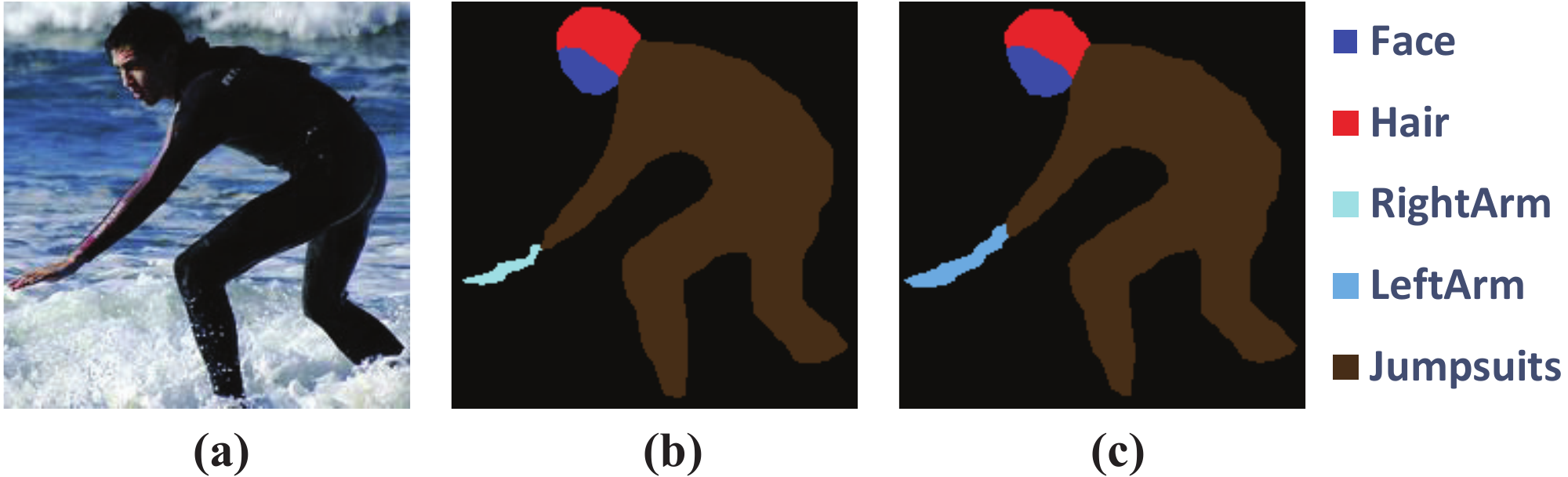}
\vspace{-3mm}
\caption{An example shows that self-supervied structure-sensitive learning is helpful for human parsing. (a): The original image. (b): The parsing results by Attention-to-scale~\cite{chen2015attention} where the left-arm is wrongly labeled as right-arm. (c): Our parsing results successfully incorporate the structure information to generate reasonable outputs.}
\label{fig: example_compare}
\vspace{-6mm}
\end{figure}

The recent progress on human parsing~\cite{chen2015attention,xia2015zoom,yamaguchi2012parsing,Yamaguchiparsing13,Dongparsing13,SimoSerraACCV2014,M-CNN,Co-CNN} has been achieved by improving the feature representations using convolutional neural networks and recurrent neural networks. To capture rich structure information, they combine CNNs and the graphical models (e.g., Conditional Random Fields (CRFs)), similar to the general object segmentation approaches~\cite{crfasrnn,chen2014semantic,wang2015joint}. However, evaluated on the new LIP dataset, the results of some existing methods~\cite{badrinarayanan2015segnet,long2014fully,chen2014semantic,chen2015attention} are unsatisfactory. Without imposing human body structure priors, these general approaches based on bottom-up appearance information sometimes tend to produce unreasonable results (e.g., right arm connected with left shoulder), as shown in Fig.~\ref{fig: example_compare}. The human body structural information has been previously well-explored in the human pose estimation~\cite{yang2016end,Chen_NIPS14} where dense joint annotations are provided. However, since human parsing requires more extensive and detailed prediction than pose estimation, it is difficult to directly utilize joint-based pose estimation models in pixel-wise prediction to incorporate the complex structure constraints. In order to explicitly enforce the produced parsing results to be semantically consistent with the human pose / joint structures, we propose a novel structure-sensitive learning approach for human parsing. In addition to using the traditional pixel-wise part annotations as the supervision, we introduce a structure-sensitive loss to evaluate the quality of predicted parsing results from a joint structure perspective. That means a satisfactory parsing result should be able to preserve a reasonable joint structure (e.g., the spatial layouts of human parts). Note that annotating both pixel-wise labeling map and pose joints is expensive and may cause ambiguities. Therefore in this work we generate approximated human joints directly from the parsing annotations and use them as the supervision signal for the structure-sensitive loss, which is hence called a  ``self-supervised" strategy, noted as Self-supervised Structure-sensitive Learning (SSL).

Our  contributions are summarized in the following three aspects. 1) We propose a new large-scale benchmark and an evaluation server to advance the human parsing research, in which 50,462 images with pixel-wise annotations on 19 semantic part labels are provided. 2) By experimenting on our benchmark, we present the detailed analyses about the existing human parsing approaches to gain some insights into the success and failures of these approaches. 3) We propose a novel self-supervised structure-sensitive learning framework for human parsing, which is capable of explicitly enforcing the consistency between the parsing results and the human joint structures. Our proposed framework significantly surpasses the previous methods on both the existing PASCAL-Person-Part dataset~\cite{chen2014detect} and our new LIP dataset.

\begin{table}[]
\centering
\scriptsize
\begin{tabular}{ccccc}
\toprule[0.7pt]
    Dataset                                & \#Training & \#Validation & \#Test & Categories \\ \hline 
    Fashionista~\cite{yamaguchi2012parsing}  & 456           & -           & 229       &     56   \\
    PASCAL-Person-Part~\cite{chen2014detect} & 1,716      & -            & 1,817     &     7      \\
    ATR~\cite{Co-CNN}                            & 16,000     & 700          & 1,000     &     18     \\ \hline
    LIP                                      & \textbf{30,462}     & \textbf{10,000}    & \textbf{10,000}    &   \textbf{20}      \\ 
\toprule[0.7pt]
\end{tabular}
\vspace{-4mm}
\caption{Overview of the publicly available datasets for human parsing. For each dataset we report the number of annotated persons in training, validation and test sets as well as the number of categories including background.}
\label{tab:dataset_num}
\vspace{-6mm}
\end{table}

\subsection{Related Work}
\textbf{Human parsing datasets:}
The commonly used publicly available datasets for human parsing are summarized in Table.~\ref{tab:dataset_num}. The previous datasets were labeled with limited number of images or categories. Containing 50,462 images annotated with 20 categories, our LIP dataset is the largest and most comprehensive human parsing dataset to date. Some other datasets in the vision community were dedicated to the tasks of clothes recognition, retrieval~\cite{liuLQWTcvpr16DeepFashion,WhereToBuyItICCV15} and human pose estimation~\cite{andriluka14cvpr,h36m_pami}, while our LIP dataset only focuses on human parsing.

\textbf{Human parsing approaches:}  
Recently many research efforts have been devoted to human parsing~\cite{Co-CNN,yamaguchi2012parsing,Yamaguchiparsing13,SimoSerraACCV2014,M-CNN,xia2015zoom,chen2015attention}. For example, Liang \etal~\cite{Co-CNN} proposed a novel Co-CNN architecture which integrates multiple levels of image contexts into a unified nerwork. Besides human parsing, there has also been increasing research interest on the part segmentation of other objects such as animals or cars~\cite{wang2014semantic,wang2015joint,lu2014parsing}. To capture the rich structure information based on the advanced CNN architecture, common solutions inlcude the combination of CNNs and CRFs~\cite{chen2014semantic,crfasrnn} and the adoptions of multi-scale feature representations~\cite{chen2014semantic,chen2015attention,xia2015zoom}. Chen \etal~\cite{chen2015attention} proposed an attention mechanism that learns to weight the multi-scale features at each pixel location. Some previous works~\cite{dong2014towards,xia2016pose} explored human pose information to guide human parsing by generating ``pose-guided'' part segment proposals. To leverage human joint structure more effortlessly and efficiently, the focus in our approach is nevertheless a new self-supervised structure-sensitive learning approach, which actually can be embedded in any networks.


\section{Look into Person Benchmark}
In this section we introduce our new ``Look into Person (LIP)'', a new large-scale dataset focusing on semantic understanding of human bodies which has several appealing properties. First, with 50,462 annotated images, LIP is an order of magnitude larger and more challenging than previous similar attempts\cite{yamaguchi2012parsing,chen2014detect,Co-CNN}. Second, LIP is annotated with elaborated pixel-wise annotations with 19 semantic human part labels and one background label. Third, the images collected from the real-world scenarios contain people appearing with challenging poses and viewpoints, heavy occlusions, various appearances and in wide range of resolutions. Furthermore, the background of images in the LIP dataset is also more complex and diverse than the one in previous datasets. Some examples are showed in Fig.~\ref{fig:dataset_example}. With the LIP dataset, we propose a new benchmark suite for human parsing together with a standard evaluation server where the test set will be kept secret to avoid overfitting.
\subsection{Image Annotation}
The images in the LIP dataset are cropped person instances from Microsoft COCO~\cite{DBLP:journals/corr/LinMBHPRDZ14} training and validation sets. We defined 19 human parts or clothes labels for annotation, which are hat, hair, sunglasses, upper-clothes, dress, coat, socks, pants, gloves, scarf, skirt, jumpsuits, face, right arm, left arm, right leg, left leg, right shoe, left shoe, and in addition to a background label. We implemented an annotation tool and generated multi-scale superpixels of images based on~\cite{amfm_pami2011} to speed up the annotation.
\subsection{Dataset splits}
In total, there are 50,462 images in the LIP dataset including 19,081 full-body images, 13,672 upper-body images, 403 lower-body images, 3,386 head-missed images, 2,778 back-view images and 21,028 images with occlusions. We split the images into separate training, validation and test sets. Following random selection, we arrive at a unique split consisting of 30,462 training and 10,000 validation images with publicly available annotations, as well as 10,000 test images with annotations withheld for benchmarking purpose. 

\begin{figure}[]
\centering
   \includegraphics[width=0.9\linewidth]{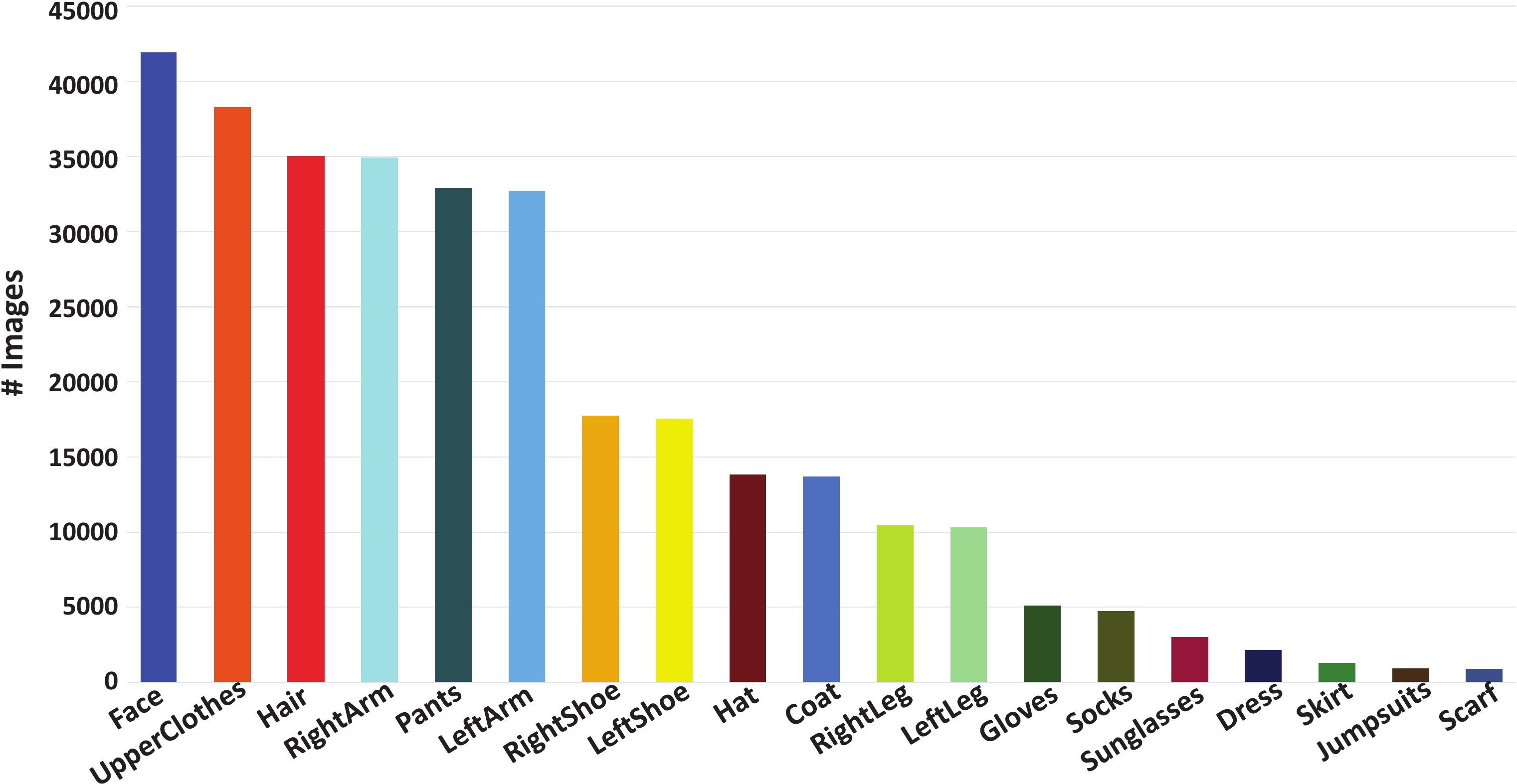}
\vspace{-4mm}
\caption{The data distribution on 19 semantic part labels in the LIP dataset.}
\vspace{-4mm}
\label{fig: dataset_analysis_label}
\end{figure}
\begin{figure}[]
\centering
   \includegraphics[width=0.8\linewidth]{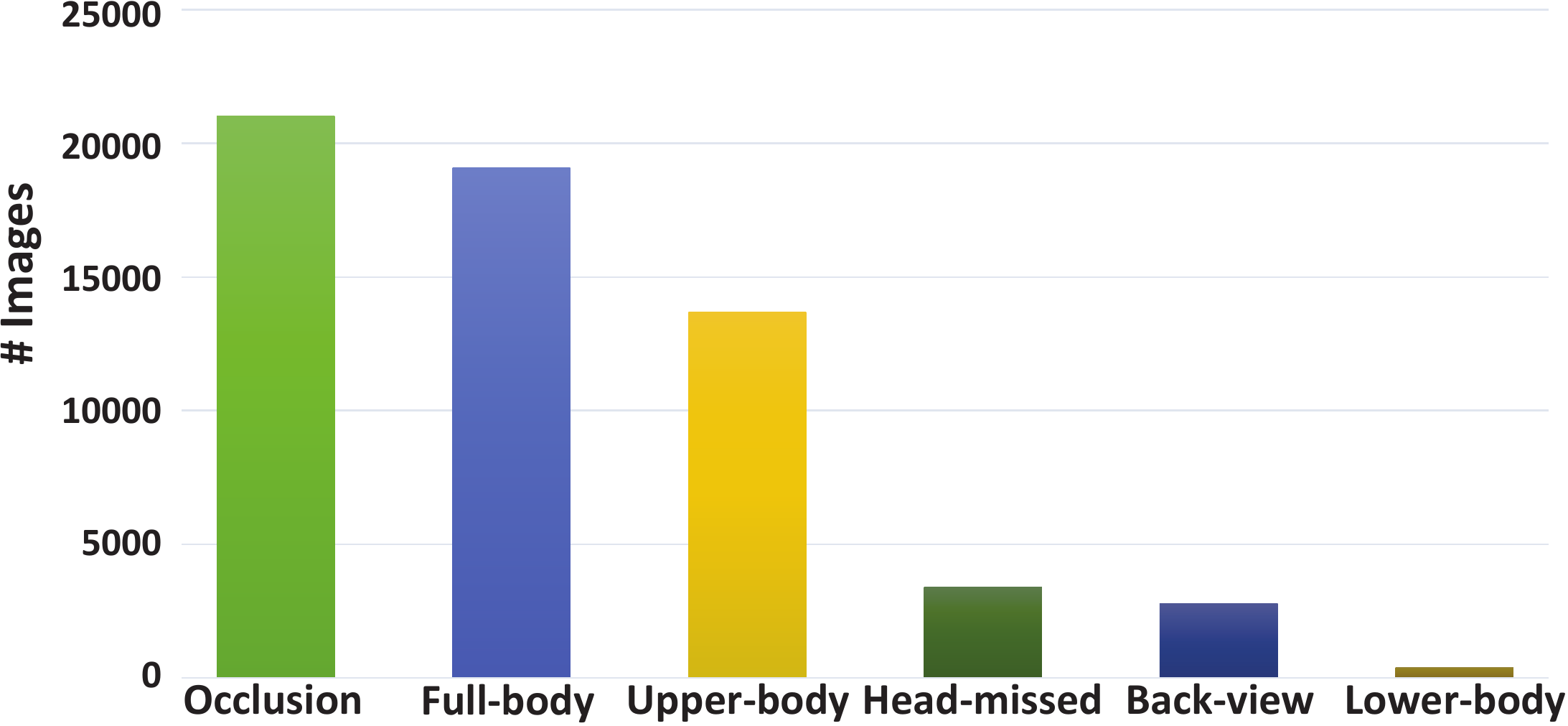}
\vspace{-3mm}
\caption{The numbers of images that show diverse visibilities in the LIP dataset, including occlusion, full-body, upper-body, lower-body, head-missed and back-view.}
\vspace{-6mm}
\label{fig: dataset_analysis_factor}
\end{figure}
\begin{figure}[]
\centering
   \includegraphics[width=0.85\linewidth]{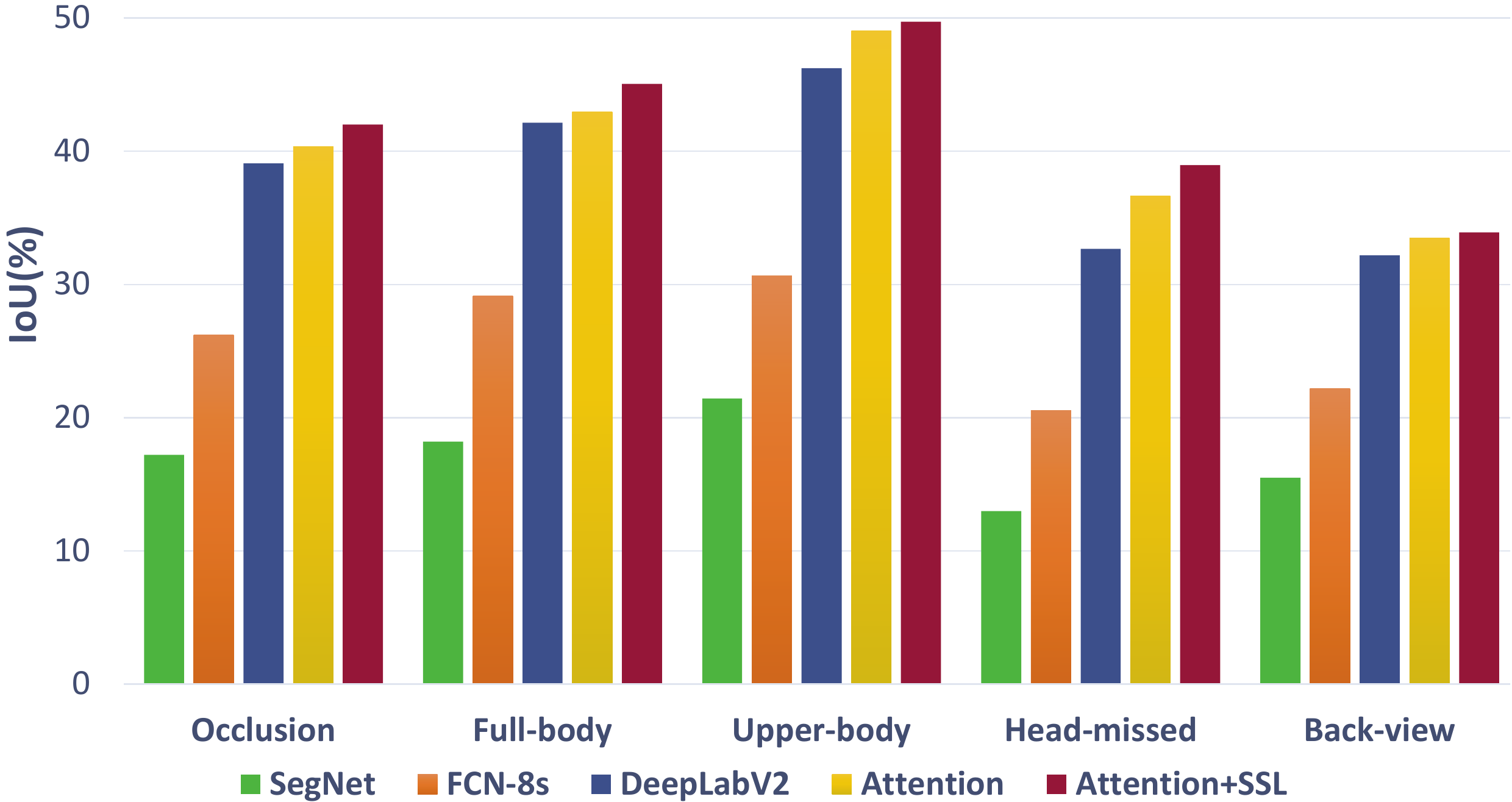}
\vspace{-4mm}
\caption{Performance comparison evaluated on the LIP validation set with different appearance, including occlusion, full-body, upper-body, head-missed and back-view.}
\vspace{-6mm}
\label{fig: analysis_val}
\end{figure}

\subsection{Dataset statistics}
In this section we analyse the images and categories in the LIP dataset in detail. In general, face, arms and legs are the most remarkable parts of a human body. However, human parsing aims to analyse every detailed regions of a person including different body parts as well as different categories of clothes. We therefore define 6 body parts and 13 clothes categories. Among these 6 body parts, we divide arms and legs into left side and right side for more precise analysis, which also increases the difficulty of the task. As for clothes classes, we have not only common clothes like upper clothes, pants and shoes, but also infrequent categories such as skirts and jumpsuits. Furthermore, small scale accessories like sunglasses, gloves and socks are also taken into account. The numbers of images for each semantic part label are presented in Fig.~\ref{fig: dataset_analysis_label}

The images in the LIP dataset contain diverse human appearances, viewpoints and occlusions. Additionally, more than half of the images suffer occlusions of different degrees. Occlusion is considered occurred if any of the 19 semantic parts appears in the image but is occluded or invisible. In more challenging cases, the images contain person instances in a back-view, which gives rise to more ambiguity of left and right spatial layouts. The numbers of images of different appearance (i.e.\ occlusion, full-body, upper-body, head-missed, back-view and lower-body) are summarized in Fig.~\ref{fig: dataset_analysis_factor}.

\section{Empirical study of state-of-the-arts}
In this section we analyse the performance of leading human parsing or semantic object segmentation approaches on our benchmark. We take advantage of our rich annotations and conduct a detailed analysis of various factors influencing the results, such as appearance, foreshortening and viewpoints. The goal of this analysis is to evaluate the robustness of the current approaches in various challenges for human parsing, and identify the existing limitations to stimulate further research advances.

\begin{table}[]
\centering
\scriptsize
\begin{tabular}{cccc}
\toprule[0.7pt]
   Method                                 & Overall accuracy    & Mean accuracy     & Mean IoU  \\ \hline 
   SegNet~\cite{badrinarayanan2015segnet}       & 69.04            & 24.00           & 18.17    \\
   FCN-8s~\cite{long2014fully}                  & 76.06            & 36.75           & 28.29    \\
   DeepLabV2~\cite{chen2014semantic}     & 82.66            & 51.64           & 41.64     \\
   Attention~\cite{chen2015attention}  & 83.43            & 54.39           & 42.92     \\ \hline
   DeepLabV2 + SSL                       & 83.16            & 52.55           & 42.44     \\
   Attention + SSL                     & \textbf{84.36}   & \textbf{54.94}  & \textbf{44.73}    \\
\toprule[0.7pt]
\end{tabular}
\vspace{-4mm}
\caption{Comparison of human parsing performance with four state-of-the-art methods on the LIP validation set.}
\vspace{-4mm}
\label{tab: lip_val}
\end{table}
\begin{table}[]
\centering
\scriptsize
\begin{tabular}{cccc}
\toprule[0.7pt]
   Method                                       & Overall accuracy & Mean accuracy   & Mean IoU  \\ \hline 
   SegNet~\cite{badrinarayanan2015segnet}       & 69.10            & 24.26           & 18.37    \\
   FCN-8s~\cite{long2014fully}                  & 76.28            & 37.18           & 28.69    \\
   DeepLabV2~\cite{chen2014semantic}     & 82.89            & 51.53           & 41.56     \\
   Attention~\cite{chen2015attention}  & 83.56            & 54.28           & 42.97     \\ \hline
   DeepLabV2 + SSL                       & 83.37            & 52.53           & 42.46     \\
   Attention + SSL                     & \textbf{84.53}   & \textbf{54.81}  & \textbf{44.59}    \\
\toprule[0.7pt]
\end{tabular}
\vspace{-4mm}
\caption{Comparison of human parsing performance with four state-of-the-art methods on the LIP test set.}
\vspace{-6mm}
\label{tab: lip_test}
\end{table}

\begin{table*}[t]
\centering
\scriptsize
\tabcolsep 0.015in 
\begin{tabular}{cccccccccccccccccccccc}
\toprule[0.7pt]
Method                                          & hat & hair & gloves & sunglasses & u-clothes & dress & coat & socks & pants & jumpsuits & scarf & skirt & face & l-arm 
                                                 & r-arm & l-leg & r-leg & l-shoe & r-shoe & Bkg & Avg\\ \hline 
SegNet~\cite{badrinarayanan2015segnet}          & 26.60 & 44.01 & 0.01  & 0.00  & 34.46 & 0.00  & 15.97 & 3.59  & 33.56 & 0.01  & 0.00  & 0.00  & 52.38 & 15.30 & 24.23
                                                & 13.82 & 13.17 & 9.26  & 6.47  & 70.62 & 18.17  \\
FCN-8s~\cite{long2014fully}                     & 39.79 & 58.96 & 5.32  & 3.08  & 49.08 & 12.36 & 26.82 & 15.66 & 49.41 & 6.48  & 0.00  & 2.16  & 62.65 & 29.78 & 36.63
                                                & 28.12 & 26.05 & 17.76 & 17.70 & 78.02 & 28.29  \\
DeepLabV2~\cite{chen2014semantic}               & 57.94 & 66.11 & 28.50 & 18.40 & 60.94 & 23.17 & 47.03 & 34.51 & 64.00 & 22.38 & 14.29 & 18.74 & 69.70 & 49.44 & 51.66
                                                & 37.49 & 34.60 & \textbf{28.22} & 22.41 & 83.25 & 41.64    \\
Attention~\cite{chen2015attention}     & 58.87 & 66.78 & 23.32 & 19.48 & 63.20 & \textbf{29.63} & 49.70 & 35.23 & 66.04 & \textbf{24.73} & 12.84 & 20.41 & 70.58
                                                & 50.17 & 54.03 & 38.35 & 37.70 & 26.20 & 27.09 & 84.00 & 42.92   \\ \hline
DeepLabV2 + SSL                                 & 58.41 & 66.22 & 28.76 & 20.05 & 62.26 & 21.18 & 48.17 & 36.12 & 65.16 & 22.94 & 14.84 & 19.37 & 70.01 & 50.45 & 53.39 
                                                & 37.59 & 36.96 & 26.29 & 26.87 & 83.67 & 42.44     \\                                                
Attention + SSL                                 & \textbf{59.75} & \textbf{67.25} & \textbf{28.95} & \textbf{21.57} & \textbf{65.30} & 29.49 & \textbf{51.92} 
                                                & \textbf{38.52} & \textbf{68.02} & 24.48 & \textbf{14.92} & \textbf{24.32} & \textbf{71.01} & \textbf{52.64} 
                                                & \textbf{55.79} & \textbf{40.23} & \textbf{38.80} & 28.08 & \textbf{29.03} & \textbf{84.56} & \textbf{44.73}  \\
\toprule[0.7pt]
\end{tabular}
\vspace{-4mm}
\caption{Performance comparison in terms of per-class IoU with four state-of-the-art methods on LIP validation set.}
\vspace{-6mm}
\label{tab: val_detail}
\end{table*}

In our analysis, we consider fully convolutional networks~\cite{long2014fully} (FCN-8s), a deep convolutional encoder-decoder architecture~\cite{badrinarayanan2015segnet} (SegNet), deep convolutional nets with atrous convolution and multi-scale~\cite{chen2014semantic} (DeepLabV2) and an attention mechanism~\cite{chen2015attention} (Attention), which all achieved excellent performance on semantic image segmentations in different ways and have completely available codes. For a fair comparison, we train each method on our LIP training set for 30 epochs and evaluate on the validation set and the test set. For DeepLabV2, we use the VGG-16 model without dense CRFs. Following~\cite{chen2015attention,xia2015zoom}, we use the standard intersection over union (IoU) criterion and pixel-wise accuracy for evaluation.
\subsection{Overall performance evaluation}
We begin our analysis by reporting the overall human parsing performance of each approach and summarize the results in Table.~\ref{tab: lip_val} and Table.~\ref{tab: lip_test}. On the LIP validation set, among the four approaches, Attention~\cite{chen2015attention} achieves the best result of 54.39\% mean accuracy, benefited from the attention model that softly weights the multi-scale features. For mean IoU, Attention~\cite{chen2015attention} performs best with 42.92\%, while both FCN-8s~\cite{long2014fully} (28.29\%) and SegNet~\cite{badrinarayanan2015segnet} (18.17\%) perform significantly worse. Similar performance is observed on the LIP test set. The interesting outcome of this comparison is that the achieved performance is substantially lower than the current best results on other segmentation benchmark such as PASCAL VOC~\cite{everingham2012pascal}. This suggests that detailed human parsing due to the small parts and diverse fine-grained labels, is more challenging than object-level segmentation, which deserves more attention in the future.

\subsection{Performance evaluation under different challenges}
We further analyse the performance of each approach with respect to the following five challenging factors: occlusion, full-body, upper-body, head-missed and back-view (see Fig.~\ref{fig: analysis_val}). We evaluate the above four approaches on the LIP validation set which contains 4,277 images with occlusions, 5,452 full-body images, 793 upper-body images, 112 head-missed images and 661 back-view images. As expected, the performance varies when affected by different factors. Back-view is clearly the most challenging case. For example, the IoU of Attention~\cite{chen2015attention} drops from 42.92\% to 33.50\%. The second most influential factor is the appearance of head. The scores of all approaches are much lower on head-missed images than the average score on the whole set. The performance also suffers a lot from occlusion. And the results of full-body images are the closest to the average level. By contrast, upper-body is relatively the easiest case, where fewer semantic parts are present and the part regions are usually larger. From the results, we can draw a conclusion that head(or face) is an important cue for the existing human parsing approaches. The probability of ambiguous results will increase if the head part disappears in the images or in the back-view. Moreover, the parts or clothes on the lower-body are more difficult than the ones on the upper-body because of the existence of small labels such as shoes or socks. In this case, body joint structure can play an effective role in guiding human parsing.
\subsection{Per-class performance evaluation}
In order to discuss and analyse each of the 20 labels in the LIP dataset in more detail, we further report the performance of per-class IoU on the LIP validation set, shown in Table.~\ref{tab: val_detail}. We observe that the results with respect to labels with larger regions like face, upperclothes, coats and pants are much better than the ones on the small-region labels such as sunglasses, scarf and skirt. Attention~\cite{chen2015attention} and DeepLabV2~\cite{chen2014semantic} perform better on small labels thanks to the utilization of multi-scale features.
\subsection{Visualization comparison}
The qualitative comparisons of four approaches on our LIP validation set are visualized in Fig.~\ref{fig:Comparison}. We display example parsing results of the five challenging factors scenarios. For the upper-body image(a) with slight occlusion, four approaches perform well with fewer errors. For the back-view image(b), all the four methods mistakenly label the right arm as left arm. The worst results appear when it comes to the head-missed image(c). SegNet~\cite{badrinarayanan2015segnet} and FCN-8s~\cite{long2014fully} fail to recognize arms and legs, while DeepLabV2~\cite{chen2014semantic} and Attention~\cite{chen2015attention} have errors on the right and left of arms, legs and shoes. Furthermore, severe occlusion(d) also affects the performance a lot. Full-body is less challenging but the small objects in a full-body image(e) like shoes are also hard to be predicted precisely. Moreover, observed from (c) and (d), some of the results are unreasonable from the perspective of human body configuration(e.g. two shoes in one foot), because the existing approaches lack the consideration of body structures. In summary, human parsing is more difficult than the general object segmentation. Particularly, human body structures should be paid more attention to strengthen the ability to predict human parts and clothes with more reasonable configurations. As a result, we consider connecting human parsing results and body joint structure to find out a better approach for human parsing.
\begin{figure*}[t]
\centering
   \includegraphics[width=0.8\linewidth]{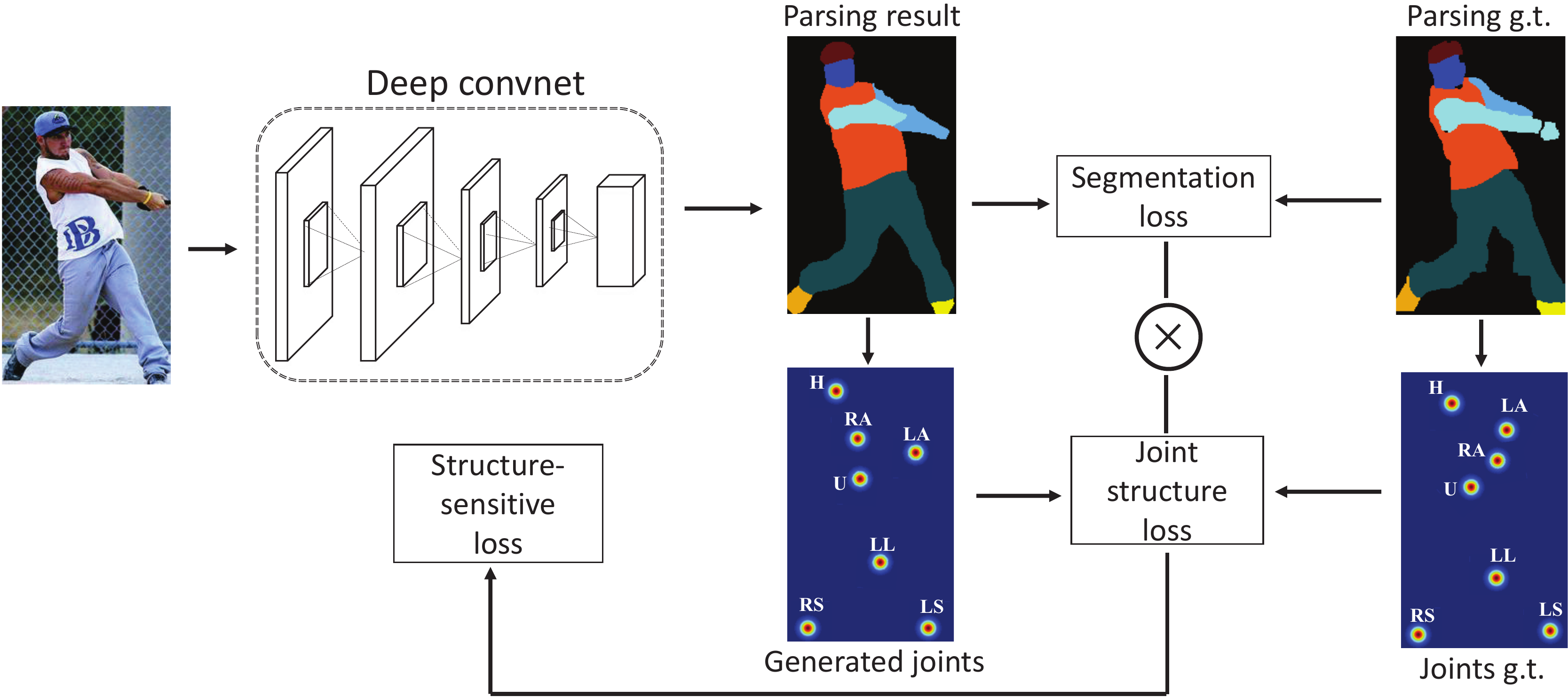}
\vspace{-2mm}
\caption{Illustration of our Self-supervised Structure-sensitive Learning for human parsing. An input image goes through parsing networks including several convolutional layers to generate the parsing results. The generated joints and joints ground truth represented as heatmaps are obtained by computing the center points of corresponding regions in parsing maps, including head (H), upper body (U), lower body (L), right arm (RA), left arm (LA), right leg (RL), left leg (LL), right shoe (RS), left shoe (LS). The structure-sensitive loss is generated by weighting segmentation loss with joint structure loss. For clear observation, here we combine nine heatmaps into one map.}
\label{fig:framework}
\vspace{-6mm}
\end{figure*}
\section{Self-supervised Structure-sensitive Learning}
\subsection{Overview}
As previously mentioned, a major limitation of the existing human parsing approaches is the lack of consideration of human body configuration, which is mainly investigated in the human pose estimation problem. The human parsing and pose estimation aim to label each image with different granularities, that is, pixel-wise semantic labeling versus joint-wise structure prediction. The pixel-wise labeling can address more detailed information while joint-wise structure provides more high-level structure. However, the results of state-of-the-art pose estimation models~\cite{yang2016end,Chen_NIPS14} still have many errors. The predicted joints do not have high enough quality to guide human parsing compared with the joints extracted from parsing annotations. Moreover, the joints in pose estimation are not aligned with parsing annotations. For example, the arms are labeled as arms for parsing annotations only if they are not covered by any clothes, while the pose annotations are independent with clothes. To address these issues, in this work, we investigate how to leverage informative high-level structure cues to guide pixel-wise prediction. We propose a novel self-supervised structure-sensitive learning for human parsing, which introduces a self-supervised structure-sensitive loss to evaluate the quality of predicted parsing results from a joint structure perspective, as illustrated in Fig.~\ref{fig:framework}.

Specifically, in addition to using the traditional pixel-wise annotations as the supervision, we generate the approximated human joints directly from the parsing annotations which can also guide human parsing training. In order to explicitly enforce the produced parsing results semantically consistent with the human joint structures, we treat the joint structure loss as a weight of segmentation loss which becomes our structure-sensitive loss.

\subsection{Self-supervised Structure-sensitive Loss}
\begin{figure}[t]
\centering
\includegraphics[width=0.9\linewidth]{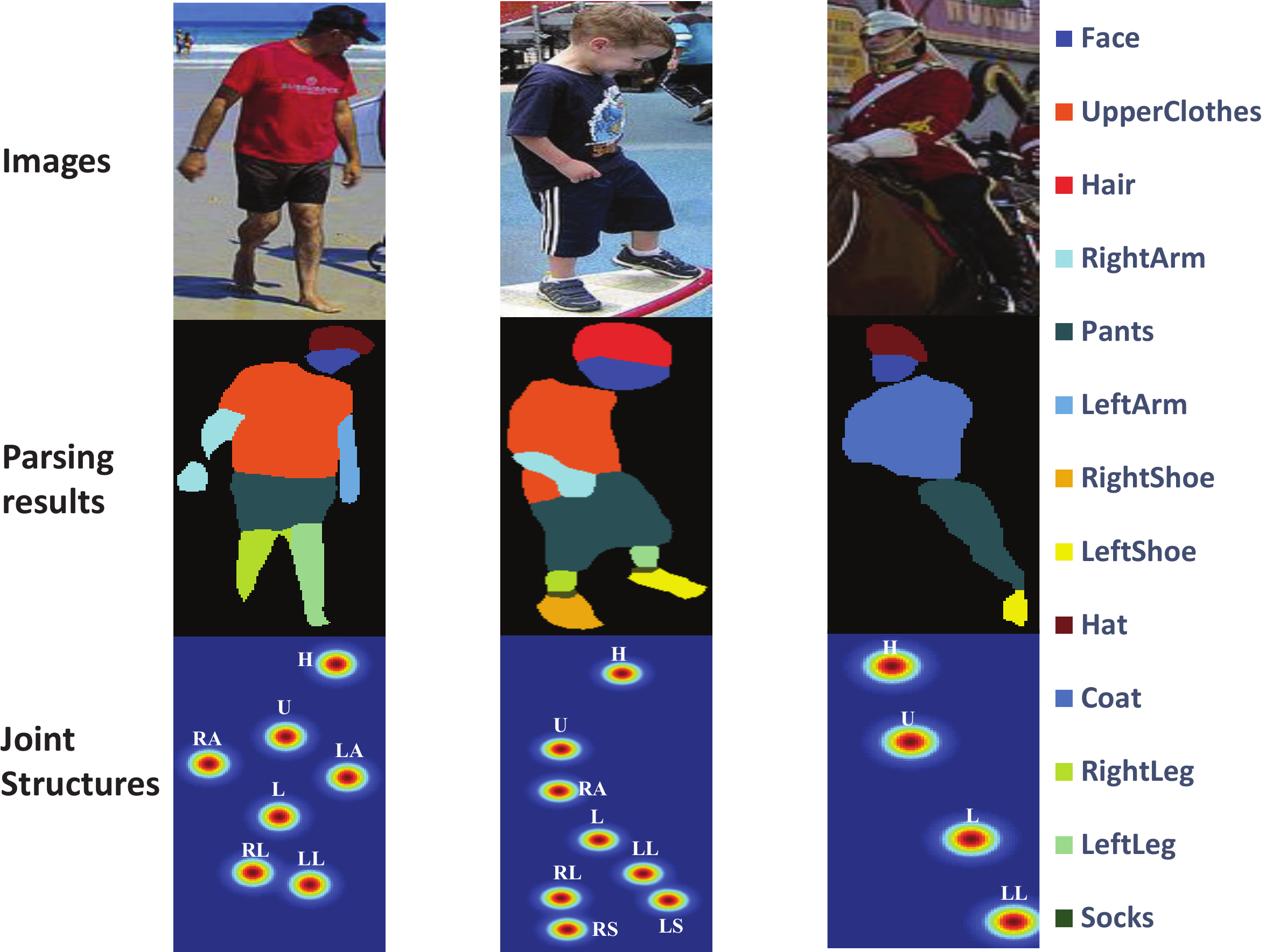}
\vspace{-3mm}
\caption{Some examples of self-supervised human joints generated from our parsing results for different bodies.}
\vspace{-4mm}
\label{fig:heatmap_example}
\end{figure}

Generally for the human parsing task, no other extensive information is provided besides the pixel-wise annotations. It means instead of using augmentative information, we have to find a structure-sensitive supervision from the parsing annotations. As the human parsing results are semantic parts with pixel-level labels, we try to explore pose information contained in human parsing results. We define 9 joints to construct a pose structure, which are the centers of regions of head, upper body, lower body, left arm, right arm, left leg, right leg, left shoe and right shoe. The region of head are generated by merging parsing labels of hat, hair, sunglasses and face. Similarly, upper-clothes, coat and scarf are merged to be upper body, pants and skirt for lower body. The rest regions can also be obtained by corresponding labels. Some examples of generated human joints for different humans are shown in Fig.~\ref{fig:heatmap_example}. Following~\cite{Pfister15a}, for each parsing result and corresponding ground truth, we compute the center points of regions to obtain joints represented as heatmaps for training more smoothly. Then we use Euclidean metric to evaluate the quality of the generated joint structures, which also reflect the structure consistency between the predicted parsing results and the ground truth. Finally, the pixel-wise segmentation loss is weighted by the joint structure loss, which becomes our structure-sensitive loss. Consequently the overall human parsing networks become self-supervised with the structure-sensitive loss. 

Formally, given an image $I$, we define a list of joints configurations $C^{P}_I = \{c^{p}_i|i\in[1,N]\}$, where $c^{p}_i$ is the heatmap of i-th joint computed according to the parsing result map. Similarly, $C^{GT}_I = \{c^{gt}_i|i\in[1,N]\}$, which is obtained from corresponding parsing ground truth. Here $N$ is a variate decided by the human bodies in the input images which equals to 9 for a full-body image. For the joints missed in the image, we simply replace the heatmaps with maps filled with zeros. The joint structure loss is the Euclidean (L2) loss, calculated as:
\vspace{-4mm}
\begin{equation}
L_{\text{Joint}} = \frac{1}{2N} \sum \limits_{i=1}^N \| c^{p}_i - c_i^{gt} \|_2^2
\end{equation}
\vspace{-1mm}
The final structure-sensitive loss, denoted as $L_{\text{structure}}$, is the combination of the joint structure loss and the parsing segmentation loss, calculated as:
\vspace{-1mm}
\begin{equation}
L_{\text{Structure}} = L_{\text{Joint}} \cdot L_{\text{Parsing}}
\end{equation}
\vspace{-1mm}
where $L_{\text{Parsing}}$ is the pixel-wise softmax loss calculated based on the parsing annotations.

We phrase our learning framework ``self-supervised'' as this above structure-sensitive loss can be generated from existing parsing results without any extra information. Our self-supervised learning framework thus has excellent adaptability and extensibility which can be injected into any advanced networks to help incorporate rich
high-level knowledge about human joints from a global perspective. 

\section{Experiments}
\subsection{Experimental Settings}
\textbf{Dataset: }
We evaluate the performance of our self-supervised structure-sensitive learning method on human parsing task on two challenging datasets. One is the public PASCAL-Person-part dataset with 1,716 images for training and 1,817 for testing, which pays attention to the human part segmentation annotated by~\cite{chen2014detect}. Following~\cite{chen2015attention,xia2015zoom}, the annotations are merge to be six person part classes and one background class which are Head, Torse, Upper / Lower arms and Upper / Lower legs. The other is our large-scale LIP dataset which is highly challenging with severe pose complexity, heavily occlusions and body truncation, as introduced and analyzed in Section 3.

\textbf{Network architecture: }
We utilize the publicly available model, Attention~\cite{chen2015attention}, as the basic architecture due to its leading accuracy and competitive efficiency. We also train a network based on DeepLabV2~\cite{chen2014semantic}, which employs re-purposed VGG-16 by atrous convolution, multi-scale inputs and atrous spatial pyramid pooling.

\textbf{Training: }
We use the pre-trained models and networks settings provided by DeepLabV2~\cite{chen2014semantic}. The scale of the input images is fiexed as $321 \times 321$ for training networks based on Attention~\cite{chen2015attention}. Two training steps are employed to train the networks. First, we train the basic network on our LIP dataset for 30 epochs, which takes about two days. Then we perform ``self-supervised" strategy to fine-tune our model with structure-sensitive loss. We fine-tune the networks for roughly 20 epochs and it takes about one and a half days. We train all the models using stochastic gradient descent with a batch size of 10 images, momentum of 0.9, and weight decay of 0.0005. In the testing stage, one images takes 0.5 second on average.

\textbf{Reproducibility:}
The proposed method is implemented by extending the Caffe framework. All networks are trained on a single NVIDIA GeForce GTX TITAN X GPU with 12GB memory. The code and
models are available at \url{https://github.com/Engineering-Course/LIP_SSL}.

\begin{table}[]
\centering
\scriptsize
\tabcolsep 0.025in 
\begin{tabular}{ccccccccc}
\toprule[0.7pt]
   Method                                     &  head  &  torso  &  u-arms &  l-arms &  u-legs &  l-legs &  Bkg   &  Avg    \\ \hline
   DeepLab-LargeFOV~\cite{chen2014semantic}      & 78.09  &  54.02  &  37.29  &  36.85  &  33.73  &  29.61  &  92.85 &  51.78  \\
   HAZN~\cite{xia2015zoom}                       & 80.79  &  59.11  &  43.05  &  42.76  &  38.99  &  34.46  &  93.59 &  56.11  \\  
   Attention~\cite{chen2015attention}   & 81.47  &  59.06  &  44.15  &  42.50  &  38.28  &  35.62  &  93.65 &  56.39  \\ 
   LG-LSTM~\cite{liang2015semantic}              & 82.72  &  60.99  &  45.40  &  \textbf{47.76}  &  \textbf{42.33}  &  37.96  &  88.63 &  57.97  \\ \hline    
   Attention + SSL                      & \textbf{83.26}  &  \textbf{62.40}  &  \textbf{47.80}  &  45.58  &  42.32 &  \textbf{39.48}  &  \textbf{94.68} 
                                        &  \textbf{59.36}   \\
\toprule[0.7pt]
\end{tabular}
\vspace{-4mm}
\caption{Comparison of person part segmentation performance with four state-of-the-art methods on the PASCAL-Person-Part dataset~\cite{chen2014detect}.}
\vspace{-6mm}
\label{tab: pascal}
\end{table}

\begin{figure*}[t]
\centering
\includegraphics[width=0.85\linewidth]{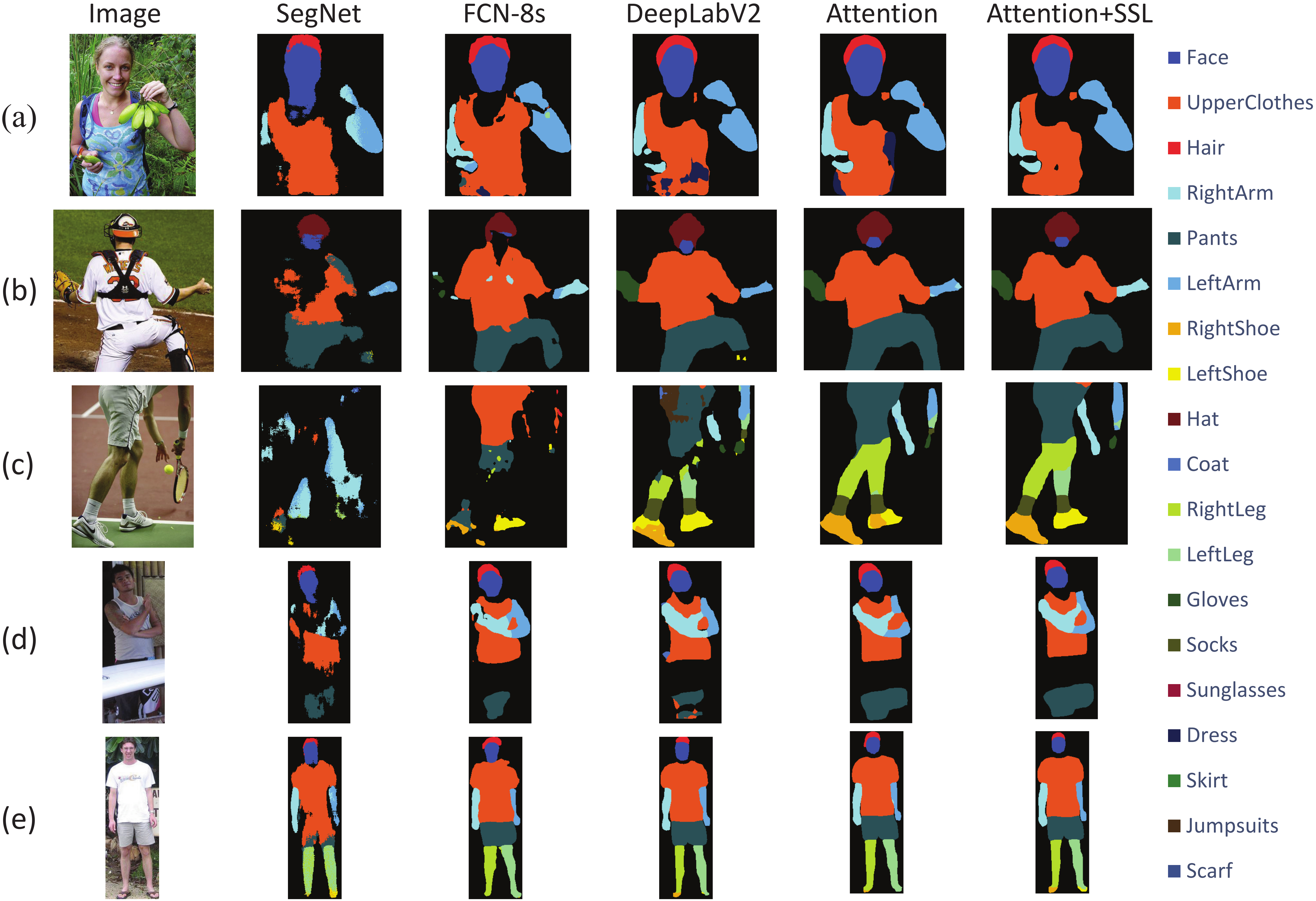}
\vspace{-4mm}
\caption{Visualized comparison of human parsing results on the LIP validation set. (a): The upper-body images. (b): The back-view images. (c): The head-missed images. (d): The images with occlusion. (e): The full-body images.}
\vspace{-6mm}
\label{fig:Comparison}
\end{figure*}

\begin{table}[]
\centering
\scriptsize
\tabcolsep 0.04in 
\begin{tabular}{c|cc|ccc|ccc}
\toprule[0.5pt]
Method                                 & ATR      & LIP       & small    & medium    & large    & 153            & 321            & 513          \\ \hline 
SegNet~\cite{badrinarayanan2015segnet} & 15.79    & 21.79     & 16.53    & 18.58     & 18.18    & 16.92          & 18.37          & 16.44        \\
FCN-8s~\cite{long2014fully}            & 34.44    & 32.28     & 22.37    & 29.41     & 28.09    & 14.52          & 15.55          & 16.25        \\ 
DeepLabV2~\cite{chen2014semantic}      & 48.64    & 43.97     & 28.77    & 40.74     & 43.02    & 36.49          & 37.59          & 37.28        \\
Attention~\cite{chen2015attention}     & 49.35    & 45.38     & 31.71    & 41.61     & 44.90    & -              & -              & -            \\ \hline
DeepLab + SSL                          & 49.92    & 44.81     & 30.05    & 41.50     & 44.10    & \textbf{38.27} & \textbf{38.97} & \textbf{39.84}  \\
Attention + SSL                        & \textbf{52.69} & \textbf{46.85} & \textbf{33.48} & \textbf{43.12} & \textbf{46.73} & -   & -      & -    \\
\toprule[0.5pt]
\end{tabular}
\vspace{-4mm}
\caption{Performance comparison in terms of mean IoU. Left: different test sets. Middle: different sizes of objects. Right: different single input sizes.}
\vspace{-6mm}
\label{tab: lip_size}
\end{table}

\subsection{Results and Comparisons}
We compare the proposed method with the strong baselines on the two public dataset.

\textbf{PASCAL-Person-Part dataset~\cite{chen2014detect}.}
Table.~\ref{tab: pascal} shows the performance of our models and comparisons with four state-of-the-art methods on the standard intersection over union (IoU) criterion. Our method can significantly outperform four baselines, particularly. For example, our best model achieves 59.36\%, 7.58\% better than DeepLab-LargeFOV~\cite{chen2014semantic} and 2.97\% better than Attention~\cite{chen2015attention}. This large improvement demonstrates that our self-supervised strategy is significantly helpful for human parsing task.

\textbf{LIP dataset:}
We report the results and the comparisons with four state-of-the-art methods on LIP validation set and test set in Table.~\ref{tab: lip_val} and Table.~\ref{tab: lip_test}. On validation set, the proposed architecture can give a huge boost in average IoU: 3.09\% better than DeepLabV2~\cite{chen2014semantic} and 1.81\% better than Attention~\cite{chen2015attention}. On test set, our method also outperforms other baselines. This superior performance achieved by our method demonstrates the effectiveness of our self-supervised structure-sensitive learning, which incorporates the body joint structure into the pixel-wise prediction.

In Fig.~\ref{fig: analysis_val}, we show the results with respect to the different challenging factors on our LIP validation set. With our structure-sensitive loss, the performance of all kinds of types are improved, which demonstrates that human joint structure is helpful for human parsing task and our self-supervised learning is reasonable and efficient.

We further report per-class IoU on LIP validation set to verify the detailed effectiveness of our structure-sensitive loss, presented in Table.~\ref{tab: val_detail}. With structure-sensitive loss, we achieved the best performance on almost all the classes. As observed from the reported results, structure-sensitive loss significantly improves the performance of the labels like arms, legs, and shoes, which demonstrates its ability to refine the ambiguous of left and right. Furthermore, the labels covering small regions such as sunglasses, socks, gloves, are predicted better with higher IoU. This improvement also demonstrates the effectiveness of structure-sensitive loss especially for small labels.

\subsection{Qualitative Comparison}
The qualitative comparisons of parsing results on the LIP validation set are visualized in Fig.~\ref{fig:Comparison}. As can be observed from these visualized comparisons, our self-learning structure outputs more semantically meaningful and precise predictions than other four methods despite the existence of large appearance and position variations. For example, observed from the full-body image(e), the small regions (e.g. left or right shoe) can be successfully segmented out by our method with structure-sensitive loss. Taking (b) and (c) for example, our approach can also successfully handle the confusing labels such as left arm versus right arm and left leg versus right leg. These regions with similar appearances can be recognized and separated by the guidance from joint structure information. For the most difficult head-missed image(c), the left shoe, right shoe and part of the left leg are excellently corrected by our approach. In general, by effectively exploiting self-supervised structure-sensitive loss, our approach outputs more reasonable results for confusing labels on the human parsing task.  

\subsection{Further experiments and analyses}
For a better understanding of our LIP dataset, we evaluate the models trained on LIP and test on ATR~\cite{Co-CNN} on the common categories, as reported in Table~\ref{tab: lip_size} (Left). In general, the performance on ATR are better than those on LIP because LIP dataset contains the instances with more diverse poses, appearance patterns, occlusions and resolution issues, which is more consistent with real-world situations. 

Following MSCOCO dataset~\cite{DBLP:journals/corr/LinMBHPRDZ14}, we have done an empirical analysis on different object sizes, i.e., small ($area < 153^2$), medium ($153^2 \leq area < 321^2$) and large ($area \geq 321^2$). The results of four baselines and the proposed SSL are reported in Table~\ref{tab: lip_size} (Middle). It can be observed that our SSL shows substantial superior performance for different sizes of objects. It further demonstrates the advantage of incorporating the structure-sensitive loss into parsing model.

To further research the influence of the input size, we perform an experiment over the scale of single input. The detailed analyses over different input sizes for all methods (except Attention~\cite{chen2015attention} for its attention mechanism over scales) are presented in Table~\ref{tab: lip_size} (Right), which shows that our structure-sensitive learning is more robust for input size.

\section{Conclusions}
In this work, we presented ``Look into Person (LIP)", a large-scale human parsing dataset and a carefully designed benchmark to spark progress in human parsing. LIP contains 50,462 images, which are richly labeled with 19 semantic part labels. Taking advantage of our rich annotations, we performed detailed experimental analyses to identify the success and limitations of the leading human parsing approaches. Furthermore, we design a novel learning strategy, namely self-supervised structure-sensitive learning, to explicitly enforce the produced parsing results semantically consistent with the human joint structures. 


{\small
\bibliographystyle{ieee}
\bibliography{egbib}
}

\end{document}